\pdfoutput=1

\documentclass[11pt]{article}
\usepackage[table]{xcolor}
\usepackage[final]{acl}

\usepackage{times}
\usepackage{latexsym}

\usepackage[T1]{fontenc}

\usepackage[utf8]{inputenc}

\usepackage{microtype}

\usepackage{inconsolata}

\usepackage{graphicx}

\usepackage{times}
\usepackage{soul}
\usepackage{url}
\usepackage{amsmath}
\usepackage{amsthm}
\usepackage{booktabs}
\usepackage{algorithm}
\usepackage{algorithmic}
\usepackage[switch]{lineno}
\usepackage{multirow} 
\usepackage{subcaption} 
\usepackage{bm}
\usepackage{pifont}
\usepackage{amsmath}
\usepackage{adjustbox}
\usepackage{booktabs}   
\usepackage{siunitx}    
\usepackage{array}

\usepackage{titlesec}
\titlespacing{\section}{0pt}{*0.9}{*0.5}
\titlespacing{\subsubsection}{0pt}{*0}{*0}
\title{SwiftPrune: Hessian-Free Weight Pruning for Large Language Models}

\author{
 \textbf{Yuhan Kang\textsuperscript{1}},
 \textbf{Yang Shi\textsuperscript{1}},
 \textbf{Mei Wen\textsuperscript{1}},
 \textbf{Jun He\textsuperscript{1}},
\\
 \textbf{Jianchao Yang\textsuperscript{1}},
 \textbf{Zeyu Xue\textsuperscript{1}},
 \textbf{Jing Feng\textsuperscript{1}},
 \textbf{Xinwang Liu \textsuperscript{1}},
\\
 \textsuperscript{1}National University of Defense Technology,
\\
 \small{
   \textbf{Correspondence:} \href{mailto:email@domain}{{kangyuhan, shiyang14, meiwen, hejun19, yangjianchao16, xuezeyu18,fengjing22,xinwangliu}@nudt.edu.cn}
 }
}

\begin{document}
\maketitle
\begin{abstract}
Post-training pruning, as one of the key techniques for compressing large language models, plays a vital role in lightweight model deployment and model sparsity. However, current mainstream pruning methods dependent on the Hessian matrix face significant limitations in both pruning speed and practical effectiveness due to the computationally intensive nature of second-order derivative calculations. This paper presents SwiftPrune, a novel Hessian-free weight pruning method that achieves hardware-efficient model compression through two key innovations: 1) SwiftPrune eliminates the need for computationally intensive Hessian matrix calculations by introducing a contribution-based weight metric, which evaluates the importance of weights without relying on second-order derivatives. 2) we employ the Exponentially Weighted Moving Average (EWMA) technique to bypass weight sorting, enabling the selection of weights that contribute most to LLM accuracy and further reducing time complexity. Our approach is extended to support structured sparsity pruning, facilitating efficient execution on modern hardware accelerators. We validate the SwiftPrune on three LLMs (namely LLaMA2, LLaMA3, and Pythia), demonstrating that it significantly enhances compression performance. The experimental findings reveal that SwiftPrune completes the pruning process within seconds, achieving an average speedup of 12.29$\times$ (up to 56.02$\times$) over existing SOTA approaches.
\end{abstract}

\section{Introduction}
In recent years, the capabilities of Large Language Models (LLMs) have experienced explosive growth. However, this advancement comes at the cost of exponential expansion in model scale, resulting in significant financial and energy expenditures~\cite{zhao2023survey}. Consequently, there has been growing effort to mitigate these costs through model compression.~\cite{Frantar2022GPTQAP, lin2024awq,frantar2023sparsegpt, LLM-pruner, sun2024a, Pruner-zero}. Among these, pruning has emerged as one of the most widely adopted techniques, with its fundamental principle involving the elimination of redundant parameters by selectively zeroing out network weights.

Contemporary pruning methods for large language models primarily eliminate retraining requirements through Hessian-based loss analysis~\cite{frantar2022optimal,fang2023depgraph,frantar2023sparsegpt,sawmya2024sparse,shao2024one}. While mathematically elegant, these methods face persistent implementation challenges due to slow pruning speeds. Specifically, computing second-order derivatives across all network weights creates a Hessian matrix whose dimensionality scales quadratically with parameter count, leading to intractable computational complexity. This limitation becomes critical in emerging real-time pruning scenarios such as  training sparse models from scratch~\cite{evci2020rigging}, finding the optimal sparsity~\cite{10.5555/3600270.3603020} and other scenarios requiring frequent pruning operations\cite{Shen_2022_CVPR,NEURIPS2022_987bed99,fu2024lazyllmdynamictokenpruning,le2025probe}. With existing methods requiring hundreds of seconds per pruning iteration (see Table~\ref{methodlatency}), conventional approaches fail to meet real-time operational demands, making the development of efficient pruning algorithms imperative for practical deployment.

Furthermore, the emergence of advanced GPU architectures underscores the demand for structured hardware-aware pruning methods that achieve genuine acceleration while maintaining computational efficiency~\cite{liu2017throughput,lu2022mt,tang2022mentha,xia2024sheared}, thereby highlighting the importance of pruning approaches compatible with structured sparse formats.


In this study, we propose SwiftPrune, a novel pruning method designed to circumvent the high computational complexity associated with Hessian matrix and its inverse calculations by developing an alternative algorithm. First, our observations indicate that identifying weights with minimal loss contribution depends more on their relative importance than on absolute values. To leverage this, SwiftPrune replaces Hessian matrix computations by constructing a numerically preserved sequence as contribution-oriented weight metrics, derived through a series of loss values. Secondly, we introduce the Exponentially Weighted Moving Average (EWMA) method, borrowed from the Transmission Control Protocol (TCP), to replace traditional sorting methods, further reducing computational complexity. Moreover, we extend this approach to support structured sparsity pruning. 
\begin{table}\small
    \centering
    \begin{tabular}{ccc} 
            \toprule
            & \multicolumn{2}{c}{LLaMA2} \\ \cline{2-3} 
            Method      & 7B                       & 13B                     \\ \hline
            SparseGPT   & 410.10                   & 912.81                  \\
            Wanda       & 114.26                   & 190.02                  \\
            Pruner-Zero & 143.45                   & 165.05                  \\
            \rowcolor[HTML]{F2F2F2} 
            SwiftPrune         & \textbf{7.73}            & \textbf{16.39}                  \\ \hline
            \bottomrule
    \end{tabular}
    \caption{The time consumption (seconds) of mainstream methods.}
    \label{methodlatency}
\end{table}

We conduct comprehensive evaluations of SwiftPrune across three prominent open-source LLM families: Pythia~\cite{Biderman2023PythiaAS}, LLaMA2~\cite{Touvron2023Llama2O}, and LLaMA3~\cite{dubey2024llama}.  Compared to previous state-of-the-art methods for large language model pruning~\cite{frantar2023sparsegpt,sun2024a}, our SwiftPrune framework achieves the pruning process within seconds, delivering an average 12.29$\times$ speedup (with peak acceleration reaching 56.02$\times$) while maintaining comparable accuracy retention across standard benchmarks. Experimental results demonstrate that SwiftPrune can finish pruning tasks more rapidly without requiring any retraining or weight updates, thereby addressing application scenarios that necessitate frequent pruning.

\section{Background}
Post-training pruning has emerged as a prevalent model compression technique, originating from quantization research~\cite{opt1,opt3,opt2} and later extended to LLM pruning~\cite{sanh2020movement, kwon2022fast,fu2022depthshrinker,sun2023simple}. In neural network optimization, the primary mechanism for minimizing the target loss function involves iterative adjustment of network weights through first-order gradient computation. However, post-training pruning methods operate under a distinct paradigm: These approaches are typically applied to models that have already converged to a local (or potentially global) minimum through standard training procedures. In such optimized states, the first-order derivatives of weights with respect to the loss function asymptotically approach zero(i.e, $\displaystyle \frac{\partial E}{\partial \Delta w} \approx 0$ in equation \ref{taylor-e}). This mathematical condition fundamentally shifts the optimization focus to second-order sensitivity analysis.

To formalize this concept, we employ a Taylor expansion of the loss function E around the trained network parameters. The expansion reveals:

\begin{equation}\small
    \Delta E=\frac{\partial E}{\partial \Delta w} \Delta w+\frac{1}{2} \Delta w^{\top} \frac{\partial ^2 E}{\partial w_i \partial w_j} \Delta w+O(\Delta w^{3})
\label{taylor-e}
\end{equation}

Where higher-order terms become non-negligible precisely when first-order derivatives vanish, necessitating explicit consideration of second-order derivatives for effective post-training pruning.

From this, we can infer that if a weight has a significant second-order derivative with respect to the target function, it indicates that the convergence of the weight is not yet stable. The impact of the weight change $\delta w_i$ on the loss function is reflected by the second-order derivative $\displaystyle \frac{\partial ^ 2 E}{\partial w_i ^2} \Delta w_i$. Unfortunately, to compute the second-order derivatives of weights, we need to construct the Hessian matrix $\displaystyle H = \left[ \frac{\partial ^2 E}{\partial w_i \partial w_j} \right]$, which costs $O(N^3)$ in time complexity. Here, we use $N$ to denote the total number of weights in the model.

To characterize the differences in outputs obtained from the compressed model and the original model under the same input, we select $\displaystyle E=\sum_{i=1}^{d_{\text {row }}}\left\|w_{i} x-\widehat{w}_{i} x\right\|_{2}$ as the loss function, where $\widehat{w}_{i}$ is the $i_\mathrm{th}$ weight, and $d_\mathrm{row}$ is the dimension of a row in a module's weights. Since weights from different rows act on the same input, resulting in elements in the same column in the output, we \textbf{\textit{assume}} that for any linear layer, weights across different rows are independent of each other. Specifically, in linear layers, every row in $W$ never multiplies with another row, so there are no cross terms in loss functions, meaning they can be optimized independently. In this scenario, $H = 2XX^{\top}$.

Building upon this theoretical foundation, this work focuses on developing a novel compression approach that bypasses the high computational cost of the Hessian matrix and its inverse while closely approximating its accuracy-preserving performance, achieving a balance between runtime efficiency and precision, and enabling scalability to very large models.

\section{The SwiftPrune Method}
\subsection{Contribution-Oriented Weight Metrics}
\label{contrib-oriented-metrics}
Our objective is to identify weights that make minimal contributions to the loss function, such that their removal would not substantially affect the model's output. In this regard, our main focus lies in analyzing the relative importance of different weights rather than their absolute values, an aspect that has been largely neglected in previous research. Previous studies have evaluated the influence of individual weights on the variation of $E$  by precisely computing their contributions through the Hessian matrix. The supplementary term in the loss function is expressed as follows:

\begin{equation}
    L=\frac{1}{2} \frac{w_{q}^2}{H_{q q}^{-1}}
\end{equation}

A crucial issue arises from the fact that the matrix $\displaystyle (2 X X^{\top})$ is not positive definite, as its determinant is zero, meaning it does not possess an inverse. To address this, we introduce a small perturbation term, denoted as:

\begin{equation}
    H=2XX^{\top}+\sum_{i}\operatorname{diag}(2XX^{\top})I
\label{h-origin}
\end{equation}

Where $I$ represents the identity matrix. This ensures that matrix operations can be performed safely. When using PyTorch, numerical methods are used for matrix computation, and due to errors in floating-point calculations, $2 X X^{\top}$ can result in matrices with extremely large values, leading to instability. By incorporating these small perturbations, we achieve stability in numerical computations with almost zero overhead.

However, computing $\Delta w$ and $L$ for every weight can be computationally expensive. The time complexity of pruning primarily lies in computing the inverse matrix $H^{-1}$, which typically has a complexity of $O(n^3)$. Even with the capability to compute Hessian matrices for each row in parallel, the total time complexity remains at $\displaystyle O(n^3)+O((\frac{n}{m})^3) = O(n^3)$, where $n$ represents the number of weights in a row.

To reduce the overall time complexity, the key is to avoid the computation of $H$ and $H^{-1}$. Our goal is not to obtain the exact value of $L$ for each weight at this stage, but rather to construct a numerically stable sequence as contribution-oriented weight metrics and to derive numerical characteristics among a series of $L$ values (such as magnitudes, variance, and averages).

Denoting $\displaystyle \sum {x_i^2}$ as $S$, in Formula~\ref{h-origin} that we constructed, $\displaystyle H_{qq} = 2(x_q^2 + \frac{1}{n} S)$. Noticed that $H^*_{qq}$ is independent of $x_q$, $H_{qq}^*$ can actually be written as $\displaystyle \bm{\mathrm{det}}\left({2X_{0}{X_{0}}^{\top} + \frac{2S}{n} I}\right)$, where $X_{0}$ is the original $X$ without the $q^{\mathrm{th}}$ element. Since $\displaystyle H_{qq}^{-1} = \frac{H_{qq}^*}{\bm{\mathrm{det}}(H)}$, and

\begin{equation}
\begin{aligned}
H_{qq}^{*} = 2(\frac{2S}{n})^{n-2}(\frac{S}{n}+S-x_q^{2})
 \\
\bm{\mathrm{det}}(H) = 2(\frac{2S}{n})^{n-1}(\frac{S}{n}+S)
\end{aligned}
\end{equation}

Thus, we can express $H_{qq}^{-1}$ as:

\begin{equation}
    H_{qq}^{-1}=\frac{\displaystyle \frac{S}{n} + S - x_q^2}{\displaystyle \frac{2S}{n}(\displaystyle \frac{S}{n} + S)}  = \frac{nS+n^2(S-x_q^2)}{2S(S+nS)}
\end{equation}

Then, we can simplify further:

\begin{equation}
\begin{aligned}
    \frac{H_{qq}^{-1}}{1 - x_q^2 / S} = \frac{nS+n^2(S-x_q^2)}{2S(S+nS)} \cdot \frac{S}{S-x_q^2} \\
     = \frac{n}{2(1+n)} \cdot \frac{1}{S-x_q^2} + \frac{n^2}{2S(1+n)}
\end{aligned}
\end{equation}

In particular, in LLMs, the dimensionality parameter $n$ (e.g. 4096 in LLaMA2-7B) is large enough to ensure that the quadratic term S dominates over $x_q^2$ by orders of magnitude ($S >> x_q^2$). This significant scale disparity allows us to employ the approximation $S - x_q^2 \approx S$ with negligible error, leading to:

\begin{equation}
    \frac{H_{qq}^{-1}}{1 - x_q^2 / S} \approx \frac{n^2+n}{2S(1+n)} = C
\end{equation}

Now we observe that $\displaystyle \frac{H_{qq}^{-1}}{1 - x_q^2 / S}$ approaches a constant. Since we are concerned with the comparative magnitudes of values rather than the exact value of each $L$, we replace $H_{qq}^{-1}$ with $(1-x_q^2/S)$ to avoid computations involving the Hessian matrix. Thus, we compute $L$ as follows:

\begin{equation}
    L=\frac{1}{2} \frac{w_{q}^{2}}{1-x_{q}^{2} / S}
\label{equation-L}
\end{equation}

Where $S$ represents the sum of all $x_i^2$ for every $x_i$ in $X$.

In this formulation, the Hessian matrix is no longer needed. To determine which weights should be removed, we can simply sort the $L$ values of all weights and eliminate those with the smallest $L$ values. As we demonstrated earlier, smaller $L$ values indicate that the removal of those weights will have a minor effect on the loss function. The time complexity of computing all $L$ values is $O(n)$, while the cost of the most common sorting algorithms is $O(n \log n)$, thus reducing the overall time complexity to $O(n \log n)$.

\begin{table}\small
    \centering
    \begin{tabular}{@{}ccc@{}}
        \toprule
        \textbf{State}  & \textbf{Updating Method} & \textbf{Initial Value} \\
        \midrule
        $est$ & $ (1 - \alpha)$$est$$+ \alpha$$L_i$ & $L_0$ \\
        $dev$ & $ (1 - \beta) dev + \beta \left | \mathit{est} - \mathit{L_i} \right |$ & 0\\
        {$S$} & $ S - w_i^2$ (if pruned) & {${\sum_{i=0}^{n-1}(x_i^2)}$} \\
        \bottomrule
    \end{tabular}
    \caption{The method for tensor state update. Parameter $la$ can be tuned for different level of sparsity.}
    \label{tensor-state-update}
\end{table}

\begin{table}[]\small
    \centering
    \begin{tabular}{l *{5}{S[table-format=-1.1]}}
        \toprule
        \textbf{Param} & \multicolumn{5}{c}{\textbf{Pruning Ratio (\%)}} \\
        \cmidrule(lr){2-6}
        & 90 & 80 & 70 & 60 & 50 \\
        \midrule
        $\textbf{la}$ & -1.5 & -0.9 & -0.2 & 0.2 & 0.5 \\
        \bottomrule
    \end{tabular}
    \caption{Parameter Adjustment under Different Pruning Ratios}
    \label{tab:sparsity_adjustment}
    \vspace{0.2cm}
    \small
\end{table}

\begin{figure}[t]
	\centering
		{\includegraphics[width=0.49\textwidth]{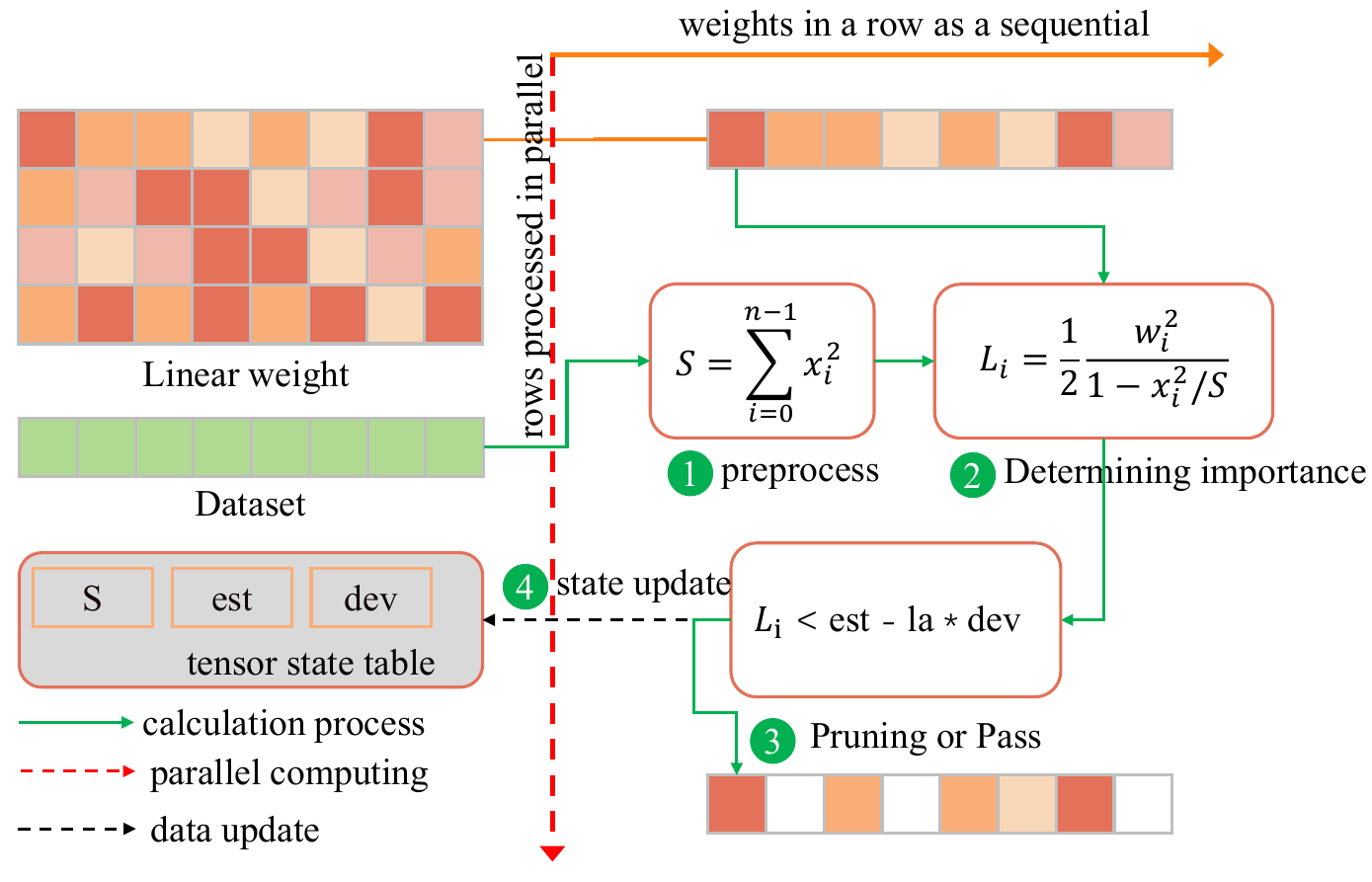}}
		\caption{Our design of novel pruning method, using EWMA criteria.}
		\label{hybrid-arch}
\end{figure}

\textbf{The Complete Algorithm.} Finally, we present the full pseudocode for SwiftPrune in Algorithm 1, including the optimizations discussed above.
\begin{algorithm}[t]\small
    \caption{The SwiftPrune algorithm. We prune the matrix $W$ to $sp$\% sparsity }
    \label{alg}
    \textbf{Input}: $W_{nrow \times ncol}$, $X_{1 \times n}$, $sp$

    \textbf{Parameter}: $\alpha,  \beta , \mathrm{la}$ 
    
    \textbf{Output}: $C_{nrow \times ncol}$
    \begin{algorithmic}[1] 
        \STATE Let $S={\sum_{i=0}^{n-1}(x_i^2)}, dev=0$ 
        \STATE \textbf{Parallel calculation for each row}
        \FOR{$i=0, 1,...,n-1$}
        \STATE $L_i=\frac{1}{2} \frac{w_{i}^{2}}{1-x_{i}^{2} / S}$
        \IF {$ i==0 $} 
        \STATE $est= L_0$
        \ENDIF
        \IF {$ L_i < \mathrm{est} - \mathrm{la} \times \mathrm{dev} $}
        \STATE $S = S - w_{i}^{2}$
        \STATE $w_i = 0$ \quad\quad\quad\quad\quad\quad\quad\quad//Pruning
        \ELSE
        \STATE \textbf{pass}
        \ENDIF
        \STATE $est = (1 - \alpha)est + \alpha \,L_i$
        \STATE $dev = (1 - \beta)dev + \beta \left | {est - L_i} \right |$
        \STATE $c_i = w_i$
        \ENDFOR
        \STATE \textbf{return} $C_{nrow \times ncol}$
    \end{algorithmic}
\end{algorithm}

\subsection{EWMA Adaption}
\label{ewma-adaption}
To further reduce the time complexity, our next objective is to find an alternative method to replace sorting, allowing us to assess where a particular $L$ value stands among all $L$ values.

The Exponentially Weighted Moving Average (EWMA) is a technique used for estimating the mean and variance of a sequence of data points. In the context of Transmission Control Protocol (TCP), it is employed to estimate the round-trip time (RTT) of a connection~\cite{paxson2011computing}.

In the practical implementation of TCP, the EWMA method exhibits strong adaptability by dynamically estimating the mean and L1-mean norm error of the recent RTT over time. We apply this method to evaluate $L$. For each row, we treat the weights as a sequential list. 

First, after calculating $S$ as outlined in Step 1 of Figure~\ref{hybrid-arch} (Algorithm~\ref{alg}, line 1), we initialize a tensor state for each weight in a row. This tensor state consists of the following components: the dynamically updated $S$, the estimated mean (denoted as $\mathrm{est}$), and the L1 mean norm error (denoted as $\mathrm{dev}$). Subsequently, following Step 2 of Figure~\ref{hybrid-arch} (Algorithm~\ref{alg}, line 4), we sequentially compute a series of $ L_i $ values. If $ L_i $ satisfies the condition $L < \mathrm{est} - \mathrm{la} \times \mathrm{dev}$ (Algorithm~\ref{alg}, line 8. The corresponding relationship between parameter $la$ and sparsity is shown in Table~\ref{tab:sparsity_adjustment}), we consider its contribution to the loss function to be minimal and prune it; otherwise, we get the original weights, as shown in Step 3 of Figure~\ref{hybrid-arch} (Algorithm~\ref{alg}, lines 9 and 12).

Next, we update the tensor state according to the procedure outlined in Table~\ref{tensor-state-update}, as illustrated in Step 4 of Figure~\ref{hybrid-arch} (Algorithm~\ref{alg}, lines 10, 14 and 16), until all weights in the row are compressed. Throughout this process, the overall time complexity is reduced to $ O(n) $. This indicates that we can evaluate the contribution of each weight to the loss function and prune the model into a sparse one within linear time.

\subsection{Structured Sparse Support}
In practical deployment scenarios, weight sparsity in large language models serves as a critical determinant for enhancing inference efficiency~\cite{tang2022mentha,liu2023deja}. To fully leverage the sparse computation capabilities of modern hardware accelerators, we extend SwiftPrune with structured sparsity support. Taking the widely adopted 2:4 fine-grained structured sparsity pattern as a representative example — a hardware-native sparse specification requiring exactly two non-zero values within every contiguous four-weight block — this design achieves deep integration with the sparse tensor computation units in NVIDIA Ampere architecture's Tensor Cores. Through instruction-level sparse format optimization, it completely eliminates format conversion overhead inherent in conventional sparsification approaches.

In implementation, we adopt fine-grained selection to support the 2:4 structured sparsity pattern. By leveraging Tensor Cores' native support for this pattern, we partition each row of weights into groups of four and identify the two smallest weights in each group through five-way comparison on average. This approach maintains the time complexity of pruning at $O(n)$ while achieving weight structured sparsity without introducing additional overhead. Compared to unstructured sparsity baselines, our method achieves 1.48$\times$ mean speedup in end-to-end inference latency (see Table~\ref{tab:sparsity_speedup}). The regularity of the sparse pattern also reduces DRAM access conflicts by 41\%, as validated through Nsight Compute memory trace analysis.

Notably, our mtehod naturally extends to 4:8 and coarser-grained structured sparsity configurations while maintaining hardware compatibility. This adaptability demonstrates our method's scalability across varying sparsity ratios without modifying the core acceleration mechanism.

\begin{table}\small
\centering
\begin{tabular}{lccc}
\toprule
\textbf{LLaMA Layer} & \textbf{Dense} & \textbf{2:4} & \textbf{Speedup} \\ 
\midrule
attn & 165.09 & 110.80 & 1.49$\times$ \\ 
attn\_qkv & 75.40 & 51.64 & 1.46$\times$ \\ 
mlp & 13.58 & 9.24 & 1.47$\times$ \\ 
\bottomrule
\end{tabular}
\caption{Comparison of inference latency(ms) between using original weights and 2:4 sparse weights for Llama2-7B on an RTX 4090 GPU}
\label{tab:sparsity_speedup}
\end{table}



\begin{table*}[ht]  \center \scriptsize
\begin{tabular}{cc|c|c|cccccccc}
\toprule 
\midrule 
\textbf{Pruning Ratio} & \textbf{Method} & \textbf{Latency(s)↓} & \textbf{WikiText2↓} & \textbf{ARC\_c} & \textbf{ARC\_e} & \textbf{WG} & \textbf{PIQA} & \textbf{HS} & \textbf{OQ} & \textbf{Avg↑}\\ 
\midrule
Dense & \textbf{LLaMA2-7B}& \_ &9.36 & 43.51 & 71.54 & 70.48 & 78.94 & 76.13 & 44.00 & 64.10 \\
\midrule
\multirow
{4}{*}{50\%} 
& magnitude & 2.29 & 44.37 & 36.77 & 53.78 & 59.74 & 70.73 & 60.88 & 36.20 & 53.01 \\
& SparseGPT& 361.29 & \textbf{7.91} & 39.33 & 66.65 & \textbf{66.61} & 76.44 & 68.84 & \textbf{39.40} & \textbf{59.54} \\
& Wanda & 108.96 & 8.01 & \textbf{39.59} & 64.85 & 65.90 & \textbf{76.61} & \textbf{69.96} & 38.40 & 59.21  \\
\rowcolor[HTML]{F2F2F2}
& SwiftPrune (ours) &  \textbf{7.85} & 8.23 & 38.40 & \textbf{67.32} & 65.27 & 75.14 & 67.18 & 38.90 & 58.70 \\ 
\midrule 
\multirow
{4}{*}{50\%(2:4)} 
& magnitude & 14.99 & 120.90 & 30.12 & 48.86 & 59.58 & 68.77 & 56.30 & 34.01 & 49.60 \\
& SparseGPT& 410.10 & \textbf{17.30} & 32.34 & 53.57 & 63.93 & 69.21 & 55.64 & 34.80 & 51.58 \\
& Wanda & 114.26 & 20.49 & 30.55 & 53.45 & 62.19 & 70.35 & 56.17 & \textbf{35.40} & 51.35  \\
\rowcolor[HTML]{F2F2F2}
& SwiftPrune (ours) &  \textbf{7.73} & 18.21 & \textbf{32.42} & \textbf{56.48} & \textbf{64.01} & \textbf{71.00} & \textbf{61.72} & 34.60 & \textbf{53.37} \\ 
\midrule
Dense & \textbf{LLaMA2-13B} & \_ &8.04 & 48.98 & 76.94 & 71.74 & 80.41 & 79.57 & 45.40 & 67.17 \\
\midrule
\multirow
{3}{*}{50\%} 
& SparseGPT & 759.13 & \textbf{9.82} & 43.60 & 69.53 & 70.88 & 78.35 & 75.13 & 44.00 & 63.58 \\
& Wanda & 146.63 & 10.03 & \textbf{46.76} & 72.85 & \textbf{71.03} & 77.71 & 76.12 & \textbf{45.60} & \textbf{65.01}  \\
\rowcolor[HTML]{F2F2F2}
& SwiftPrune (ours) &  \textbf{16.60} & 10.27 & 45.73 & \textbf{73.74} & 69.38 & \textbf{78.43} & \textbf{76.29} & 42.60 & 64.36 \\ 
\midrule 
\multirow
{3}{*}{50\%(2:4)} 
& SparseGPT & 912.81 & 13.27 & 38.65 & 66.62 & 68.67 & 73.83 & 64.54 & 41.00 & 58.88 \\
& Wanda & 190.02 & 15.61 & 37.71 & 65.49 & 66.77 & 75.41 & 62.65 & 39.00 & 57.83  \\
& SwiftPrune (ours) &  \textbf{16.39} & \textbf{9.42} & \textbf{42.30} & \textbf{75.72} & \textbf{70.40} & \textbf{79.38} & \textbf{77.28} & \textbf{45.20} & \textbf{65.04} \\ 
\midrule
Dense & \textbf{LLaMA3.1-8B} & \_ &7.93 & 53.50 & 81.10 & 73.56 & 81.23 & 78.90 & 44.80 & 68.84 \\
\midrule
\multirow
{3}{*}{50\%} 
& SparseGPT & 558.57 & 12.54 & 43.26 & 67.34 & 70.09 & 76.82 & 68.90 & 40.60 & 61.16 \\
& Wanda & 99.98 & 11.26 & \textbf{45.73} & \textbf{69.61} & 69.77 & 76.88 & \textbf{71.39} & 43.20 & \textbf{62.76}  \\
\rowcolor[HTML]{F2F2F2}
& SwiftPrune (ours) &  \textbf{9.49} & \textbf{10.96} & 44.70 & 68.10 & \textbf{70.24} & \textbf{77.25} & 70.31 & \textbf{43.82} & 62.40 \\ 
\midrule 
\multirow
{3}{*}{50\%(2:4)} 
& SparseGPT & 613.13 & 17.76 & 33.96 & 57.24 & 63.46 & 69.42 & 55.22 & 33.60 & 52.15 \\
& Wanda & 125.64 & 29.95 & 29.95 & 52.15 & 59.27 & 67.85 & 48.69 & 31.40 & 48.21  \\
\rowcolor[HTML]{F2F2F2}
& SwiftPrune (ours) &  \textbf{9.28} & \textbf{15.02} & \textbf{35.27} & \textbf{59.19} & \textbf{65.84} & \textbf{73.17} & \textbf{63.10} & \textbf{35.02} & \textbf{55.26} \\ 
\midrule 
Dense & \textbf{Pythia-2.8B} & \_ &12.69 & 32.76 & 59.01 & 58.17 & 74.10 & 59.41 & 35.00 & 53.07 \\
\midrule
\multirow
{3}{*}{50\%} 
& SparseGPT & 185.17 &  22.53 &  \textbf{29.44} & 51.58 & 56.51 & 69.31 & 50.22 & 30.80 & 47.97 \\
& Wanda & 39.77 & 23.30 & 28.16 & 49.07 & 56.04 & 68.93 & 51.01 & 30.80 & 47.33  \\
\rowcolor[HTML]{F2F2F2}
&SwiftPrune (ours)& \textbf{3.99} &\textbf{21.69} &29.18 & \textbf{51.94} & \textbf{57.22} & \textbf{70.29} & \textbf{52.14} & \textbf{31.20} & \textbf{48.66}\\
\midrule 
\multirow
{3}{*}{50\%(2:4)} 
& SparseGPT & 196.38 & 27.12 & 24.83 & \textbf{46.89} & 54.06 & 65.61 & 40.88 & 28.20  & 43.41 \\
& Wanda & 48.23 & 30.69 & 24.15 & 38.89 & 53.75 & 61.43 & 36.81 & 28.40 & 40.57  \\
\rowcolor[HTML]{F2F2F2}
&SwiftPrune (ours)&\textbf{3.83} &\textbf{23.30} &\textbf{27.28} &46.81 &\textbf{56.12} &\textbf{69.83} &\textbf{49.13} &\textbf{29.20} & \textbf{46.39}\\
\midrule 
\bottomrule
\end{tabular}
\caption{Zero-shot performance of the pruned LLaMA2-7B, LLaMA2-13B, LLaMA3.1-8B and Pythia-2.8B. “Latency(s)” indicates represents the time overhead required for overall model pruning (excluding communication time such as loading to GPU). The 'Avg' denotes the average value calculated across six classification datasets (HS, WG, and OQ represent HellaSwag, WinoGrande, and OpenbookQA respectively). Bold formatting indicates the best performance under equivalent compression ratios. However, note that for Latency(s), it represents the best performance excluding the cost associated with magnitude. The magnitude pruning method is omitted for LLaMA2-13B, LLaMA3.1-8B, and Pythia-2.8B because it causes significant accuracy degradation in these models.}
\label{ppl}
\end{table*}

\section{Experiments}
\subsection{Experimental setup}
\label{exp-setup}
\textbf{Models.} We conduct comprehensive evaluations of SwiftPrune across three prominent open-source LLM families: Pythia, LLaMA2, and LLaMA3. As a GPT-NeoX variant specialized for interpretability analysis in autoregressive transformers, Pythia provides granular architectural insights through its controlled design. The LLaMA series represents cutting-edge pre-trained models, with LLaMA3 introducing enhanced multilingual tokenization and dynamic sparse attention mechanisms in its latest iteration. This benchmark suite spans 7B to 70B parameter scales, covering interpretability-oriented frameworks, production-optimized architectures, and next-generation multilingual models, thereby systematically validating our method's robustness across evolving transformer paradigms.

\textbf{Datasets.} We evaluate pruning using zero-shot perplexity (PPL) on WikiText2~\cite{wiki}. For task-agnostic performance, we adopt LLaMA's evaluation approach, testing on OpenCompass~\cite{2023opencompass} and Lm-evaluation-harness~\cite{eval-harness} benchmarks. These benchmarks offer a comprehensive assessment for LLMs. The datasets encompassed in this assessment are as follows: ARC(Easy and Challenge)~\cite{arc}, WinoGrande~\cite{winogrande}, PIQA~\cite{bisk2020piqa}, HellaSwag~\cite{zellers2019hellaswag} and OpenbookQA~\cite{openqa}. 

\textbf{Platforms.} Our experimental platform configuration consists of 2$\times$ Intel(R) Xeon(R) Platinum 8358 CPUs @ 2.60GHz and 8$\times$ RTX 4090 GPUs (24GB VRAM each). The software stack includes GCC 7.5.0, NVIDIA CUDA 12.1, and Python 3.11.5 (Anaconda 23.9.0). For structured sparsity implementation, we utilize PyTorch 2.3.0.dev20240220+cu121 with custom kernel extensions that leverage the native 2:4 sparse tensor core operations on the RTX 4090 GPUs, enabled via the cuSPARSELt library. This implementation directly accesses the hardware's structured sparsity acceleration units, where the 2:4 compressed sparse blocks are processed through dedicated warp-level MMA (Matrix Multiply-Accumulate) instructions (SM\_AMPERE\_SPARSE\_MMA feature).

\subsection{Evaluation of SwiftPrune Algorithm}

\begin{table*}[htbp]\small
\centering
\begin{tabular}{lcccc}
\toprule
\textbf{Method} & \textbf{Weight Update} & \textbf{Calibration Data} & \textbf{Pruning Metric } $S_{ij}$ & \textbf{Complexity} \\
\midrule
Magnitude       &  {\color[HTML]{D8383A} NO}           & {\color[HTML]{D8383A} NO}              & $|W_{ij}|$                        & $O(1)$              \\
SparseGPT       & {\color[HTML]{96C37D} YES}           & {\color[HTML]{96C37D} YES}             & $\big[|W|^2 \big/ \text{diag}\big[(XX^T + \lambda I)^{-1}\big]\big]_{ij}$ & $O(d_{\text{hidden}}^3)$ \\
Wanda           & {\color[HTML]{D8383A} NO}              & {\color[HTML]{96C37D} YES}              & $|W_{ij}| \cdot \|X_j\|_2$          & $O(d_{\text{hidden}}^2)$ \\
\rowcolor[HTML]{F2F2F2} 
SwiftPrune           & {\color[HTML]{D8383A} NO}                & {\color[HTML]{96C37D} YES}             & $|W_{ij}| \cdot n$          & $O(d_{\text{hidden}})$ \\
\bottomrule
\end{tabular}
\caption{Taxonomy of Pruning Methodologies: Algorithmic Properties and Computational Complexity}
\label{tab:methods}
\end{table*}

\begin{figure*}[htbp]
\centerline{\includegraphics[width=\textwidth]{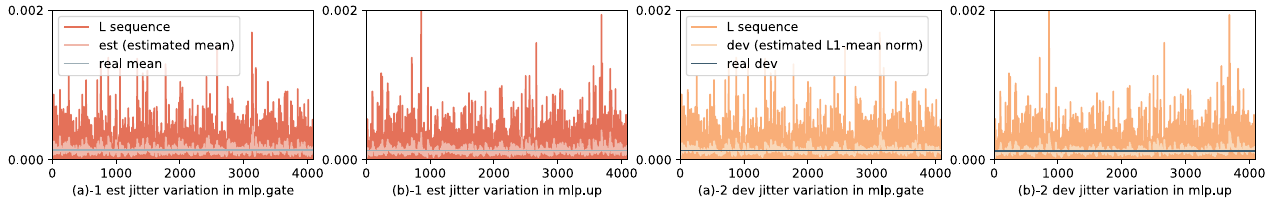}}
\caption{Statistical magnitude detection of $L$ with EWMA method in LLaMA2-7B MLP blocks. $x$ axis presents the sequence number of each weight, and $y$ axis presents the numerical values. Ideal algorithms should show est approaches real mean and dev approaches real dev.}
\label{fig-detect-change}
\vspace{-1em}
\end{figure*}

\textbf{Efficiency: The SwiftPrune algorithm provides a significant speedup.} The performance gains derive primarily from algorithmic innovations. By eliminating computationally intensive Hessian matrix calculations, our $O(n)$ algorithm achieves rapid acceleration in LLM pruning tasks without requiring retraining or weight updates (Table~\ref{tab:methods}). This methodology not only enables efficient assessment of weight significance but also maintains near-constant time complexity — a critical advantage that prevents substantial increases in computational overhead as model dimensions expand. 

Our experiments systematically demonstrate that the proposed method achieves an average speedup of 43.75$\times$ and 12.29$\times$ compared to state-of-the-art pruning approaches like SparseGPT and Wanda respectively (detailed in Table~\ref{ppl}). This substantial acceleration effectively addresses the temporal overhead inherent in scenarios requiring iterative pruning applications, particularly those involving adaptive sparsity mechanisms and dynamic input pruning techniques.

\textbf{Accuracy: Zero-shot performance comparison with baselines.} We conducted comprehensive fine-grained pruning experiments on the LLaMA2-7B model and rigorously evaluated its average zero-shot learning accuracy across six tasks under three pruning configurations (including 50\% sparsity with 2:4 structured pruning) using the lm-evaluation-harness framework. 


As shown in Table~\ref{ppl}, the experimental results demonstrate that when reaching a 50\% pruning rate, SwiftPrune maintains an average performance decline within 2 percentage points compared to the original dense model. Systematic comparative analysis reveals that our method achieves significant acceleration while preserving negligible accuracy loss (average difference < 1\%), outperforming existing approaches like Wanda and SparseGPT that rely on computationally intensive Hessian matrix calculations. Notably, in 2:4 structured sparsity scenarios, our method achieves 3.7-12.7\% accuracy improvements across multiple benchmarks through innovative fine-grained pruning strategies. These empirical findings validate the innovation of the SwiftPrune framework: It realizes intelligent model compression through algorithm-level optimizations without requiring training data, effectively balancing model performance preservation with substantial computational complexity reduction. This breakthrough provides an efficient solution for practical industrial deployment scenarios where both accuracy and processing speed are critical. Experimental results of SwiftPrune under other pruning ratio will be presented in the appendix.

\textbf{Reliability: SwiftPrune adapts to weight changes and approaches global expectations.} In Figure~\ref{fig-detect-change}, we demonstrate how our SwiftPrune method consistently and accurately predicts the mean and variations of weights. As the weight sequence lengthens, SwiftPrune exhibits improved responsiveness and faster convergence. By adjusting the smoothing factors ($\alpha$, $\beta$, and $\mathrm{la}$), we can fine-tune the algorithm's responsiveness and stability to align with specific network characteristics. This capability enables us to determine whether the current row weight significantly impacts the final output, thereby deciding whether to prune it.

The data presented in Figure~\ref{fig-detect-change}, derived from a layer of LLaMA2-7B, indicate that we can consistently approach the global weight mean shortly after an initial startup period. For the results in Figure~\ref{fig-detect-change}, we set $\alpha = 0.125$, $\beta = 0.125$, and $\mathrm{la} = 4$, which is consistent with RFC 6298 ~\cite{paxson2011computing}. This configuration remains robust even as the parameters undergo significant changes, with fluctuations staying relatively small. Our predictions consistently vary between the global variance and the global L1-mean norm, showing a pattern similar to the predicted mean. The experiments also show that the method maintains its effectiveness as the model weight length increases, showcasing high scalability and validating the feasibility of our introduced EWMA approach as a viable alternative to traditional sorting methods. We also conducted the same experiments on the Pythia-2.8B model, achieving equally strong performance and further validating the generalizability of SwiftPrune.

\textbf{Fine-tuning.} We systematically investigated two distinct fine-tuning strategies: LoRA~\cite{hu2022lora} and full-parameter dense fine-tuning~\cite{lv2023full}. Experiments were conducted on the WikiText2 training dataset while strictly maintaining the structured/unstructured mask matrices generated during pruning. We validated the compatibility of pruned models with fine-tuning algorithms under two representative sparsity patterns: unstructured 50\% sparsity and structured 2:4 sparsity.

As shown in Table~\ref{ft}, the pruned LLaMA3.1-8B model processed by SwiftPrune pruning demonstrated significant improvements in both zero-shot accuracy and perplexity metrics after fine-tuning. Experimental results confirm the strong compatibility between the adopted fine-tuning strategies and pruning methodology, effectively restoring the model's expressive power diminished during weight trimming. This finding provides crucial technical validation for efficient compression and performance preservation in LLMs.

\begin{table}[]\small\center
\begin{tabular}{ccccc}
\toprule
Evaluation                  & Dense                   & Fine-tuning                & 50\%                       & 2:4                        \\ \hline
                            &                         & \cellcolor[HTML]{F2F2F2}{\color[HTML]{D8383A} NO} & \cellcolor[HTML]{F2F2F2}62.40 & \cellcolor[HTML]{F2F2F2}55.26 \\
                            &                         & LoRA                       &    63.81                       &        58.47                  \\
\multirow{-3}{*}{Zero-Shot} & \multirow{-3}{*}{68.84} & Full                       &    66.02                       &        63.21                   \\ \hline
                            &                         & \cellcolor[HTML]{F2F2F2}{\color[HTML]{D8383A} NO} & \cellcolor[HTML]{F2F2F2}10.96 & \cellcolor[HTML]{F2F2F2}15.02 \\
                            &                         & LoRA                       &   9.53                        &         13.21                  \\
\multirow{-3}{*}{Perplexity} & \multirow{-3}{*}{7.93} & Full                       &   8.42                        &         10.32                  \\ \hline
\end{tabular}
\caption{Fine-tuning can recover some of the losses caused by pruning.}
\label{ft}
\end{table}

\section{Related Work}
The most fundamental sparsification approach is magnitude-based pruning, which achieves sparsity by setting the smallest weights to zero~\cite{han2015deep,zhu2017prune}. Although these methods scale well, they often cause significant performance degradation in LLMs~\cite{frantar2023sparsegpt,harma2024effective}. To improve sparsification, researchers learned from the Optimal Brain Surgeon (OBS) method~\cite{hassibi1993optimal}, which innovatively uses the inverse of the Hessian matrix to update unpruned weights, thereby compensating for errors caused by weight removal. However, OBS faces computational bottlenecks in practical applications - calculating and storing the inverse Hessian matrix is computationally infeasible for models with millions of parameters. To address this challenge, recent research has proposed two improvement approaches: one approximates the inverse Hessian matrix calculation, such as the WoodFisher method~\cite{singh2020woodfisher}; the other performs layerwise pruning, known as Optimal Brain Compression (OBC)~\cite{frantar2022optimal}. While these methods perform well on medium-scale networks, they struggle with larger language models~\cite{Frantar2022GPTQAP}.

SparseGPT~\cite{frantar2023sparsegpt} tackles the Hessian computation challenge through a grouping-based pruning strategy. This approach applies compensation updates to weights in adjacent columns via Hessian matrix operations while employing unstructured and semi-structured pruning patterns to streamline large language models. Concurrently, Sparse Expansion~\cite{sawmya2024sparse} enhances inference efficiency by constructing dedicated Hessian matrices for distinct input clusters, enabling specialized pruning of expert weight matrices through the SparseGPT framework. In a notable simplification, Wanda~\cite{sun2024a} demonstrates that preserving only the diagonal elements of the Hessian matrix suffices for effective pruning, significantly reducing computational overhead while maintaining competitive performance.

Simultaneously, to achieve tangible speed improvements in practical applications, there has been a growing recognition of the need to apply pruning in a structured and hardware-compatible manner~\cite{santacroce2023matters,ma2023structural,li2023losparse,xia2024sheared}. This approach is typically followed by additional training (or fine-tuning) to restore any diminished performance. For example, the LLM-pruner~\cite{ma2023llm} eliminates specific connection structures within LLMs prior to further training. Similarly, the Large Language Model Surgeon~\cite{ouderaa2024the} interleaves recovery fine-tuning with pruning.

\section{Conclusion}
\label{conclusion}
In this paper, we propose SwiftPrune, a hardware-friendly approach for pruning of LLMs. The core innovation of our study is the development of a novel Hessian-Free LLM pruning method, which significantly reduces time complexity from $O(n^3)$ to $O(n)$ compared to mainstream algorithms. This theoretical breakthrough ensures that our method consistently outperforms existing approaches in terms of both computational efficiency and scalability. Built on a rigorous mathematical foundation, SwiftPrune demonstrates exceptional effectiveness and relevance, particularly as the scale of future LLMs continues to expand. By significantly reducing computational resource demands and energy consumption.

\section{Limitations}
While this study achieves promising results and makes notable contributions to the field, we acknowledge several limitations requiring further investigation. Although our method optimizes memory usage compared to existing approaches (SwiftPrune's 20.01 GB, Wanda's 22.79 GB and SparseGPT's 30.82 GB for LLaMA2-7B), the improvements remain constrained. Consequently, further optimization of memory consumption to enable deployment of larger models on resource-constrained devices constitutes a critical focus for future research.

This study addresses critical challenges in large language model (LLM) compression, aiming to facilitate broader adoption and practical implementation of LLM technologies. In light of growing concerns regarding ethical implications associated with LLMs, particularly the potential presence of latent biases embedded within these models, we have conducted comprehensive investigations to ensure the integrity of our proposed methodology. Our findings demonstrate that the developed pruning approach not only maintains model performance but also adheres to ethical standards by preventing the amplification of existing biases or introduction of new discriminatory patterns.

\bibliography{custom}

\appendix
\section{More experimental results}

\begin{table*}[ht]  \center \scriptsize
\begin{tabular}{cc|c|c|cccccccc}
\toprule 
\midrule 
\textbf{Pruning Ratio} & \textbf{Method} & \textbf{Latency(s)↓} & \textbf{WikiText2↓} & \textbf{ARC\_c} & \textbf{ARC\_e} & \textbf{WG} & \textbf{PIQA} & \textbf{HS} & \textbf{OQ} & \textbf{Avg↑}\\ 
\midrule
Dense & LLaMA2-7B& \_ &9.36 & 43.51 & 71.54 & 70.48 & 78.94 & 76.13 & 44.00 & 64.10 \\
\midrule
\multirow
{3}{*}{10\%} 
& SparseGPT& 371.83 & 10.44 & 43.86 & 71.42 & 70.24 & \textbf{77.52} & 76.19 & 42.40 & 63.60 \\
& Wanda & 103.32 & \textbf{9.38} & \textbf{44.11} & 71.54 & \textbf{70.63} & 76.12 & 78.78 & \textbf{45.00} & \textbf{64.36} \\
\rowcolor[HTML]{F2F2F2}
& SwiftPrune (ours) &  \textbf{9.55} & 9.88 & 44.02 & \textbf{71.62} & 70.53 & 76.73 & \textbf{78.80} & 44.31 & 64.33 \\ 
\midrule
\multirow
{3}{*}{20\%} 
& SparseGPT& 371.34 & \textbf{9.56} & 43.60 & 70.79 & \textbf{69.53} & 78.29 & \textbf{76.12} & \textbf{45.20} & 63.92 \\
& Wanda & 103.51 & 9.57 & \textbf{44.03} & \textbf{71.42} & 69.29 & 78.24 & 76.04 & 44.80 & \textbf{63.97} \\
\rowcolor[HTML]{F2F2F2}
& SwiftPrune (ours) & \textbf{9.37} & 9.67 & 43.42 & 70.33 & 69.37 & \textbf{78.31} & 76.01 & 44.88 & 63.72 \\ 
\midrule
\multirow
{3}{*}{30\%} 
& SparseGPT& 357.03 & \textbf{9.86} & 43.68 & 70.07 & 69.13 & 78.07 & 75.17 & 44.20 & 63.38 \\
& Wanda & 100.04 & 9.90 & \textbf{44.11} & \textbf{70.37} & \textbf{69.29} & \textbf{78.29} & \textbf{75.30} & \textbf{45.00} & \textbf{63.72} \\
\rowcolor[HTML]{F2F2F2}
& SwiftPrune (ours) &  \textbf{9.12} & 10.02 & 43.91 & 70.01 & 69.11 & 78.04 & 75.22 & 45.01 & 63.55 \\ 
\midrule
\multirow
{3}{*}{40\%} 
& SparseGPT& 357.03 & \textbf{9.39} & \textbf{43.83} & \textbf{69.69} & \textbf{69.13} & \textbf{78.84} & 73.15 & \textbf{45.40} & \textbf{63.34} \\
& Wanda & 100.04 & 10.55 & 42.75 & 69.14 & 68.74 & 77.91 & 73.55 & 43.00 & 62.51 \\
\rowcolor[HTML]{F2F2F2}
& SwiftPrune (ours) &  \textbf{8.62} & 10.34 & 42.84 & 69.01 & 68.09 & 78.01 & \textbf{73.91} & 42.69 & 62.43 \\ 
\midrule
\bottomrule
\end{tabular}
\caption{Zero-shot performance of the pruned LLaMA2-7B. “Pruning Ratio” refers to the proportion of parameters removed relative to the original number of parameters. “Latency(s)” indicates represents the time overhead required for overall model pruning (excluding communication time such as loading to GPU). The 'Avg' denotes the average value calculated across six classification datasets (HS, WG, and OQ represent HellaSwag, WinoGrande, and OpenbookQA respectively). Bold formatting indicates the best performance under equivalent compression ratios.}
\label{ppl1}
\end{table*}

\end{document}